\documentclass[letterpaper, 10 pt, conference]{ieeeconf}  

\IEEEoverridecommandlockouts                              

\title{\LARGE \bf
DexEMG: Towards Dexterous Teleoperation System via EMG2Pose Generalization
}

\author{
Qianyou Zhao$^{1,2}$,
Wenqiao Li$^{1,2}$,
Chiyu Wang$^{1}$,
Kaifeng Zhang$^{1}$\\
$^{1}$Sharpa
$^{2}$Shanghai Jiao Tong University}

\usepackage{graphicx}
\usepackage{multirow}
\usepackage{booktabs}  
\usepackage{graphicx}
\usepackage{caption}
\usepackage{amsmath}

\begin{document}



\twocolumn[{
\renewcommand\twocolumn[1][]{#1}
\maketitle
\begin{center}
  \includegraphics[width=\textwidth]{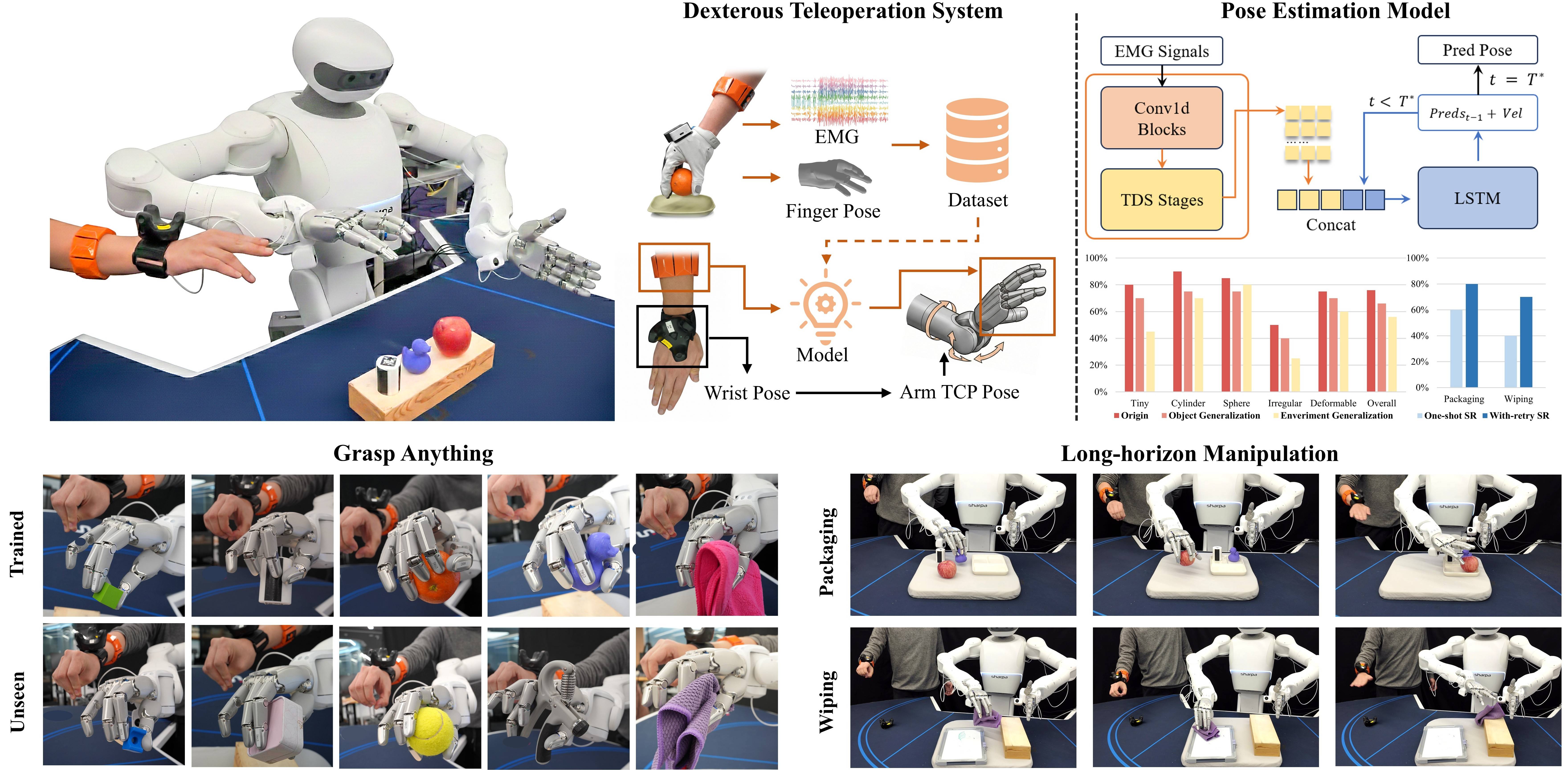}
  \captionof{figure}{Overview of the \textbf{DexEMG} system. (Top-Left) The teleoperation interface leverages sEMG signals for portable control. (Top-Right) The EMG2Pose architecture utilizes a TDS-based encoder and a feedback-enabled LSTM decoder for continuous pose regression. (Bottom) DexEMG demonstrates robust generalization across diverse object categories and successfully completes long-horizon manipulation tasks including desktop packaging and wiping.}
  \label{fig:teaser}
\vspace{5mm}
\end{center}
}]

\thispagestyle{empty}
\pagestyle{empty}
\begin{abstract}

High-fidelity teleoperation of dexterous robotic hands is essential for bringing robots into unstructured domestic environments. However, existing teleoperation systems often face a trade-off between performance and portability: vision-based capture systems are constrained by costs and line-of-sight requirements, while mechanical exoskeletons are bulky and physically restrictive. In this paper, we present DexEMG, a lightweight and cost-effective teleoperation system leveraging surface electromyography (sEMG) to bridge the gap between human intent and robotic execution. We first collect a synchronized dataset of sEMG signals and hand poses via a MoCap glove to train EMG2Pose, a neural network capable of continuously predicting hand kinematics directly from muscle activity. To ensure seamless control, we develop a robust hand retargeting algorithm that maps the predicted poses onto a multi-fingered dexterous hand in real-time. Experimental results demonstrate that DexEMG achieves high precision in diverse teleoperation tasks. Notably, our system exhibits strong generalization capabilities across novel objects and complex environments without the need for intensive individual-specific recalibration. This work offers a scalable and intuitive interface for both general-purpose robotic manipulation and assistive technologies.

\end{abstract}

\section{INTRODUCTION}
The integration of multi-fingered dexterous hands into unstructured domestic environments, such as smart homes and elderly care facilities, marks a transformative frontier in robotics \cite{1_billard2019trends}. To perform nuanced manipulation tasks in these settings, a teleoperation interface that is both highly intuitive and physically unobtrusive is indispensable \cite{3_daffertshofer2020teleoperation}. For domestic deployment, such a system must be cost-effective, easy to don and doff, and functional without the need for complex external infrastructure.

Traditionally, dexterous teleoperation has relied on two dominant paradigms, both of which face significant barriers to widespread domestic adoption. Exoskeleton-based interfaces \cite{4_perez2017wearable,5_fang2020multimodal} offer high-fidelity control but are inherently bulky and mechanically rigid. Their isomorphic design creates a hardware bottleneck—any structural modification to the robotic hand requires a corresponding, high-cost redesign of the haptic glove \cite{6_sarac2019design}. On the other hand, vision-based motion capture systems provide a non-contact alternative \cite{7_handa2020dextreme, 8_qin2022dexmv} but are often prohibitively expensive and require specialized, static infrastructure. These systems typically rely on multiple high-speed cameras and controlled environments, making them impractical for cost-sensitive and dynamic domestic applications \cite{9_antoniou2021review}.

To overcome these constraints, surface electromyography (sEMG) has emerged as a compelling modality. By capturing neuromuscular signals directly from the forearm, sEMG allows for a wearable, low-cost, and unobtrusive interface \cite{10_bi2019emg}. While prior research has successfully utilized sEMG for discrete gesture classification \cite{11_cai2021gesturedetect,12_zeng2021deep}, the transition to achieving continuous, high-dimensional pose estimation remains a formidable challenge \cite{13_quivira2018translating}.

In this work, we present DexEMG, a portable dexterous teleoperation system that utilizes a commodity sEMG wristband to bridge the gap between human intent and robotic execution. We first collected a synchronized dataset of sEMG signals and hand poses using a Manus MoCap glove to train EMG2Pose, a neural network designed for continuous hand kinematics prediction. Based on these predictions, we developed a kinematic hand retargeting algorithm to map human joint motions onto a dexterous robotic hand in real-time.

The primary contributions of this work are:

\begin{itemize}
    \item A lightweight and low-cost teleoperation system: We develop a wearable framework that eliminates the need for costly vision-based motion capture or cumbersome exoskeletons, significantly lowering the entry barrier for dexterous manipulation.
    \item We demonstrate through extensive experiments that DexEMG achieves robust performance across diverse environments, novel objects, and complex tasks, proving its efficacy for real-world deployment.
\end{itemize}

\section{RELATED WORK}

\subsection{Pose Estimation with sEMG Signals}
To achieve the high-fidelity control necessary for dexterous teleoperation, research has increasingly shifted from discrete recognition toward mapping sEMG signals directly to continuous joint kinematics. This task is inherently challenging due to the complex, non-linear, and non-stationary relationship between surface muscle activity and skeletal movement. Early regression-based approaches primarily utilized Long Short-Term Memory (LSTM) networks to account for the electromechanical delay and temporal dependencies inherent in muscle activation \cite{quivira2018translating}. Building upon these temporal models, NeuroPose \cite{wen2021neuropose} further enhanced estimation accuracy by integrating anatomical constraints into a ResNet-based architecture, effectively enabling 3D finger tracking from commodity wearable armbands while ensuring biomechanical plausibility.

More recently, the adoption of Transformer architectures has marked a significant shift in the field, leveraging self-attention mechanisms to model long-range temporal correlations more effectively than traditional Recurrent Neural Networks (RNNs) \cite{yi2021emg2pose}. These attention-based models excel at dynamically weighing relevant muscle channels, leading to superior inference efficiency and tracking precision in multi-DoF tasks. A persistent bottleneck, however, remains the significant inter-user variability of sEMG signals. To address this, the EMG2Pose framework \cite{emg2pose2024benchmark} pioneered a velocity-based regression approach trained on a massive, diverse dataset. By predicting joint velocities rather than static angles, thereby reducing sensitivity to sensor displacement, and employing meta-learning for rapid adaptation, they demonstrated robust, real-time hand tracking across hundreds of unseen users. Collectively, these advancements have laid the foundation for sEMG to serve as a viable, occlusion-free alternative to vision-based pose capture in robotic teleoperation.

\subsection{Dexterous Teleoperation Systems}
The performance of a teleoperation system is fundamentally constrained by the fidelity, latency, and robustness of the human pose capture modality \cite{nieto2014teleoperation}. Over the past decades, researchers have explored various sensing technologies, primarily categorized into mechanical-wearable and vision-based approaches.

\textbf{Mechanical and Wearable Sensing} Mechanical sensing involves physical attachments that directly measure joint kinematics. Linkage-based exoskeleton systems, such as the CyberGrasp \cite{kessler1995cybergrasp} and Dexmo \cite{gu2016dexmo}, utilize high-resolution encoders to provide precise joint angle data and integrated force feedback. However, these systems are inherently bulky and heavy, imposing a significant physical burden on the operator that leads to rapid fatigue \cite{kessler1995cybergrasp, pollard2002quantifying}. Furthermore, the complex mechanical structure makes them uncomfortable for long-term wear and extremely difficult to deploy for large-scale applications due to high maintenance and hardware costs.

As a more mobile alternative, sensorized gloves utilizing flex sensors, stretch sensors, or Inertial Measurement Units (IMUs) have been developed \cite{sundaram2019learning, fang2017pressure}. While they offer improved portability, these wearable devices suffer from sensor drift and require frequent, tedious re-calibration to adapt to varying hand anthropometry \cite{sundaram2019learning, lin2021review}, which limits their reliability in dynamic teleoperation tasks.

\textbf{Vision-based Sensing} Vision-based capture has gained popularity due to its non-intrusive and markerless nature. Professional marker-based systems (e.g., Vicon, OptiTrack) achieve sub-millimeter accuracy by tracking reflective markers with multiple infrared cameras \cite{merriaux2017study}. Despite their precision, these systems are prohibitively expensive and necessitate a strictly controlled environment with pre-installed camera arrays, rendering them unsuitable for portable or field-based teleoperation \cite{han2016remote}.

Low-cost consumer-grade sensors, such as Leap Motion and Azure Kinect, leverage depth imaging or deep-learning-based frameworks like MediaPipe to reconstruct 3D skeletons \cite{lugaresi2019mediapipe, bassily2014robots}. However, these systems are inherently susceptible to self-occlusion—a critical failure mode in dexterous manipulation where fingers are frequently hidden by the palm or grasped objects \cite{supancic2015depth, yuan2018depth}.

\begin{figure*}[t]
  \centering
  \includegraphics[width=\linewidth]{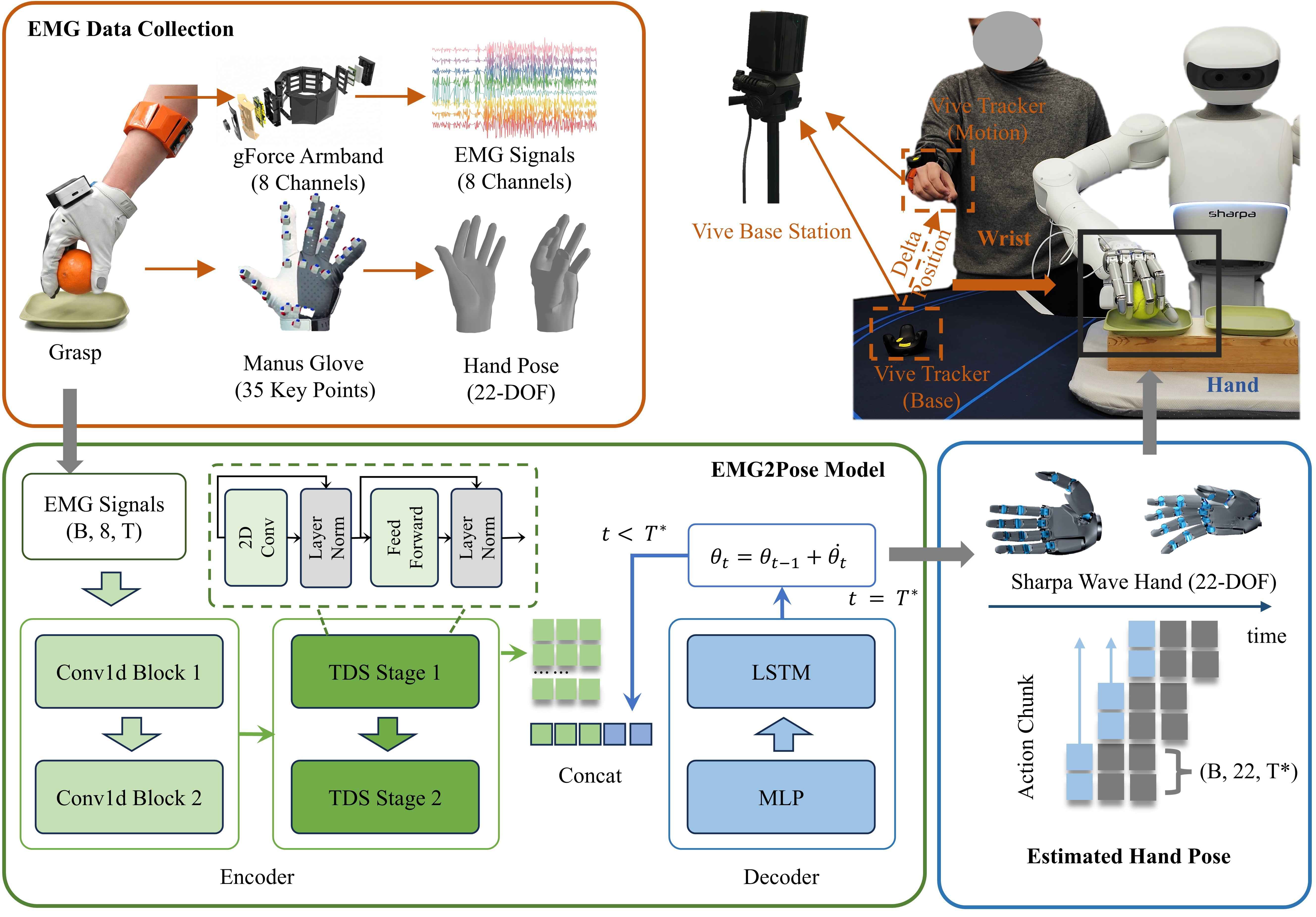}
  \caption{Overview of the DexEMG system architecture. (Top-Left) Multimodal data collection captures synchronized sEMG signals and ground-truth hand kinematics. (Bottom-Left) The EMG2Pose perception engine extracts spatial-temporal features via TDS stages and uses an LSTM-based decoder to reconstruct continuous hand poses. (Top-Right) The real-time deployment pipeline integrates the predicted hand action chunks with wrist tracking to control a multi-fingered robotic hand.}
  \label{fig:model_architecture}
  \vspace{-1mm}
\end{figure*}

In this work, we propose an sEMG-driven teleoperation system that effectively circumvents the intrinsic limitations of the aforementioned modalities. Unlike vision-based systems, sEMG signals are immune to optical occlusion, and unlike mechanical exoskeletons, our approach offers a lightweight, non-restrictive, and computationally efficient interface. By decoding muscle activity directly, our system enables high-fidelity, dexterous robotic control with minimal hardware overhead.







\section{METHOD}


The DexEMG system comprises two primary components: (1) a kinematic retargeting module that works in tandem with synchronized sEMG data collection to produce physically viable ground-truth labels, and (2) an EMG2Pose model designed to decode continuous robotic hand kinematics directly from raw muscle activity. By seamlessly integrating these components with spatial wrist tracking, we establish a robust, real-time teleoperation pipeline. Figure \ref{fig:model_architecture} illustrates the overall system architecture.

\subsection{Dataset Collection via Kinematic Retargeting}

To build a high-fidelity dataset that maps neuromuscular signals to executable robotic commands, we developed a synchronized data acquisition pipeline combining raw signal capture with kinematic retargeting.

During data collection, the operator wears a multi-channel sEMG interface to capture high-frequency muscle activation patterns from the forearm, alongside a high-precision motion capture glove that tracks the spatial trajectories of the human hand skeleton. To ensure comprehensive coverage of the dynamic workspace, the operator performs a diverse array of grasping and in-hand manipulation tasks.

We then map these captured human poses onto the robotic hand to generate geometrically consistent ground-truth joint angles. This retargeting process is formulated as a keypoint-based optimization problem, which minimizes the $L_2$ distance between corresponding anatomical landmarks (such as the fingertips and palm center) across the human and robotic workspaces:
\begin{equation}
    q^* = \arg\min_{q} \sum_{i} \| p_i^{h} - p_i^{r}(q) \|_2^2,
\end{equation}
where $q$ denotes the configuration of the robotic hand joints and $p_i$ represents the position of the $i$-th keypoint. Specifically, $p_i^{h}$ and $p_i^{r}(q)$ represent the 3D spatial coordinates of the $i$-th corresponding keypoint on the human hand and the robotic hand, respectively. By prioritizing task-relevant keypoints, this formulation ensures that the generated labels preserve the precise fingertip positioning required for fine manipulation.

To guarantee that the resulting ground-truth trajectories are physically executable, we enforce strict safety constraints during retargeting. A collision classifier, pre-trained on the robot's configuration space, evaluates the optimized joint angles $q^*$. If a potential self-collision is detected, the system dynamically clamps the pose to a safe manifold. Consequently, the final dataset pairs synchronized sEMG streams with valid, collision-free robotic joint angles, offering a robust supervision signal for subsequent model training.

\subsection{EMG2Pose Generalization}

Our teleoperation system leverages the EMG2Pose architecture~\cite{emg2pose2024benchmark} as its foundational perception engine, which maps high-dimensional sEMG signals to continuous 3D hand poses. The model follows an encoder-decoder design as shown in Fig.~\ref{fig:model_architecture} (bottom-left).

The encoder processes raw sEMG inputs of shape $(B, 8, T)$ using two 1D convolution (Conv1d) blocks followed by two Time-Depth Separable (TDS) stages. Each TDS stage comprises a 2D convolution, a layer normalization step, a feedforward layer, and a final layer normalization. The resulting feature maps, shaped $(B, 32, T^*/4)$, are resampled to a length of $T^*$ before being passed to the decoder. Rather than predicting absolute joint angles $\theta$, the decoder utilizes an LSTM network paired with a Multi-Layer Perceptron (MLP) to estimate joint velocities $\dot{\theta}$. The absolute poses are then reconstructed iteratively:
\begin{equation}
    \theta_t = \theta_{t-1} + \dot{\theta}_t,
\end{equation}
where $\theta_0$ is a predefined rest pose. This velocity-centric approach effectively decouples muscle activation intensity from static postures, which mitigates signal drift during sustained grasping and ensures smoother transitions between distinct manipulation tasks. The output is an action chunk of shape $(B, 22, T^*)$.

Because sEMG directly captures neuromuscular activity from the forearm, the resulting pose estimates are inherently immune to visual occlusions. This characteristic guarantees reliable control, even when operating in confined or cluttered workspaces.

\subsection{DexEMG System}

As illustrated in Figure \ref{fig:model_architecture}, the DexEMG pipeline unfolds across three distinct phases: data collection, model training, and real-time teleoperation.

\textbf{Data collection.}
We capture synchronized physiological and kinematic data to establish the mapping between neuromuscular activity and hand motion. The operator is equipped with a wearable multi-channel sEMG interface to record muscle activation patterns, alongside a high-fidelity motion capture system that provides ground-truth hand poses via skeletal keypoints (Fig.~\ref{fig:model_architecture}, top-left). These keypoints are subsequently re-targeted to 22-DOF joint angles, creating a rich multimodal dataset that links internal muscle effort to external hand configuration.

\textbf{Model training.}
Using the collected paired data, the EMG2Pose model is trained to regress continuous hand poses from raw sEMG streams. The velocity-based training objective provides inherent robustness to sensor displacement and signal noise across different recording sessions.

\textbf{Deployment and inference.}
The real-time inference pipeline serves as the core of our teleoperation system, enabling low-latency control of a multi-fingered robotic hand. During deployment (Fig.~\ref{fig:model_architecture}, right), the cumbersome data glove is removed; the operator relies solely on the wearable sEMG interface and a spatial tracking device. 
The system performs online inference on a sliding window of sEMG inputs, generating action chunks of predicted joint angles. By executing the initial frames of each chunk and iteratively advancing the window, the system achieves smooth, continuous pose estimation. This lightweight setup eliminates the need for external cameras, facilitating intuitive and occlusion-robust teleoperation in unconstrained workspaces.


\begin{figure*}[htb]
    \centering
    \includegraphics[width=\linewidth]{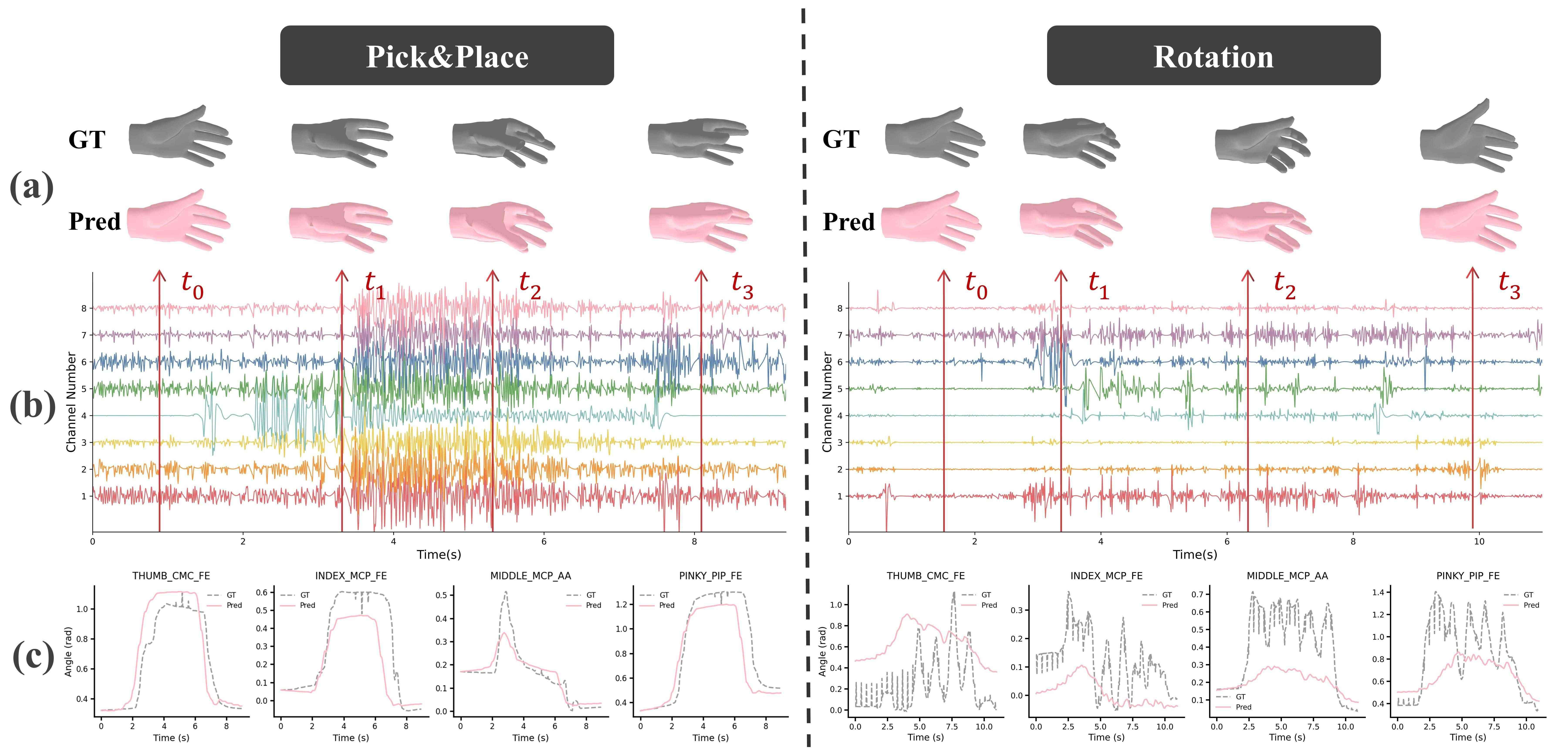}
    \caption{Representative snapshots of continuous pose estimation via EMG2Pose. EMG2Pose model accurately tracks high-dimensional hand kinematics across various grasping and in-hand manipulation gestures. The estimated poses (rendered) show high correspondence with the user's actual hand motions.}
    \label{fig:visualization}
    \vspace{-1mm}
\end{figure*}

\section{EXPERIMENT}

We conduct a series of experiments to evaluate the accuracy, generalizability, and scalability of DexEMG. The evaluation is guided by three core research questions:

Q1: Accuracy. Can the EMG2Pose network faithfully reconstruct complex hand kinematics from raw sEMG signals?

Q2: Generalizability. Does the system robustly handle novel objects and unstructured environments?

Q3: Scalability. Is the system capable of executing reliable, long-horizon manipulation sequences?

\subsection{Experimental Setup}

To evaluate the proposed DexEMG framework, we construct a hardware testbed comprising both data collection and teleoperation interfaces. 

For neuromuscular signal acquisition, the operator wears a gForce 8-channel sEMG armband. Ground-truth human hand kinematics are captured using a Manus MoCap glove, which provides 35 high-precision skeletal keypoints. During the real-time teleoperation phase, the glove is removed, and spatial wrist movements are tracked using HTC Vive Trackers. The Vive Tracker provides 6D wrist pose via a base station, with increments computed from the relative displacement between two trackers.

The target robotic platform is a Sharpa Wave hand, a multi-fingered dexterous end-effector with 22 degrees of freedom (DOF). All deep learning models are deployed on a local workstation, ensuring minimal computational latency to support continuous, high-frequency control.






\subsection{Empirical Study of Pose Estimation}

To evaluate the core capability of our EMG-based pose estimation, we visualize the performance of the EMG2Pose network across tasks with varying levels of complexity: a fundamental "Grasp" task and a highly dexterous "In-hand Rotation" task.

\textbf{Qualitative Fidelity} (Fig.~\ref{fig:visualization}-a): Fig.~\ref{fig:visualization}-a illustrates the comparison between the ground truth (GT) hand poses captured by the Manus glove and the poses predicted (Pred) by the EMG2Pose network. In the simpler grasping sequence, the predicted mesh aligns closely with the operator's actual hand posture. Notably, even in the "In-hand Rotation" task, which involves high-frequency joint coupling and subtle finger movements, the network successfully reconstructs the overall motion intention and maintains high morphological similarity.

\textbf{Signal Correlation} (Fig.~\ref{fig:visualization}-b): The multi-channel sEMG signals corresponding to different finger articulations are visualized in Fig. \ref{fig:visualization}-b. The temporal variations in the EMG signals across eight channels demonstrate distinct activation patterns for different manipulation stages (indicated by $t_0$ to $t_3$). This correlation underscores the network's ability to extract discriminative features from noisy, non-stationary muscle signals to drive the hand model.

\textbf{Quantitative Joint Tracking} (Fig.~\ref{fig:visualization}-c): To further quantify the estimation accuracy, we plotted the angular trajectories of representative key joints (e.g., THUMB CMC FE, INDEX MCP FE). As shown in Fig. \ref{fig:visualization}-c, the predicted joint angles (solid red lines) track the ground truth (dashed black lines) with minimal phase lag. While slight deviations are observed during the rapid transitions of the rotation task, the network consistently captures the dynamic range and trend of the joint movements. 
\begin{figure}[t]
  \centering
  \includegraphics[width=0.82\linewidth]{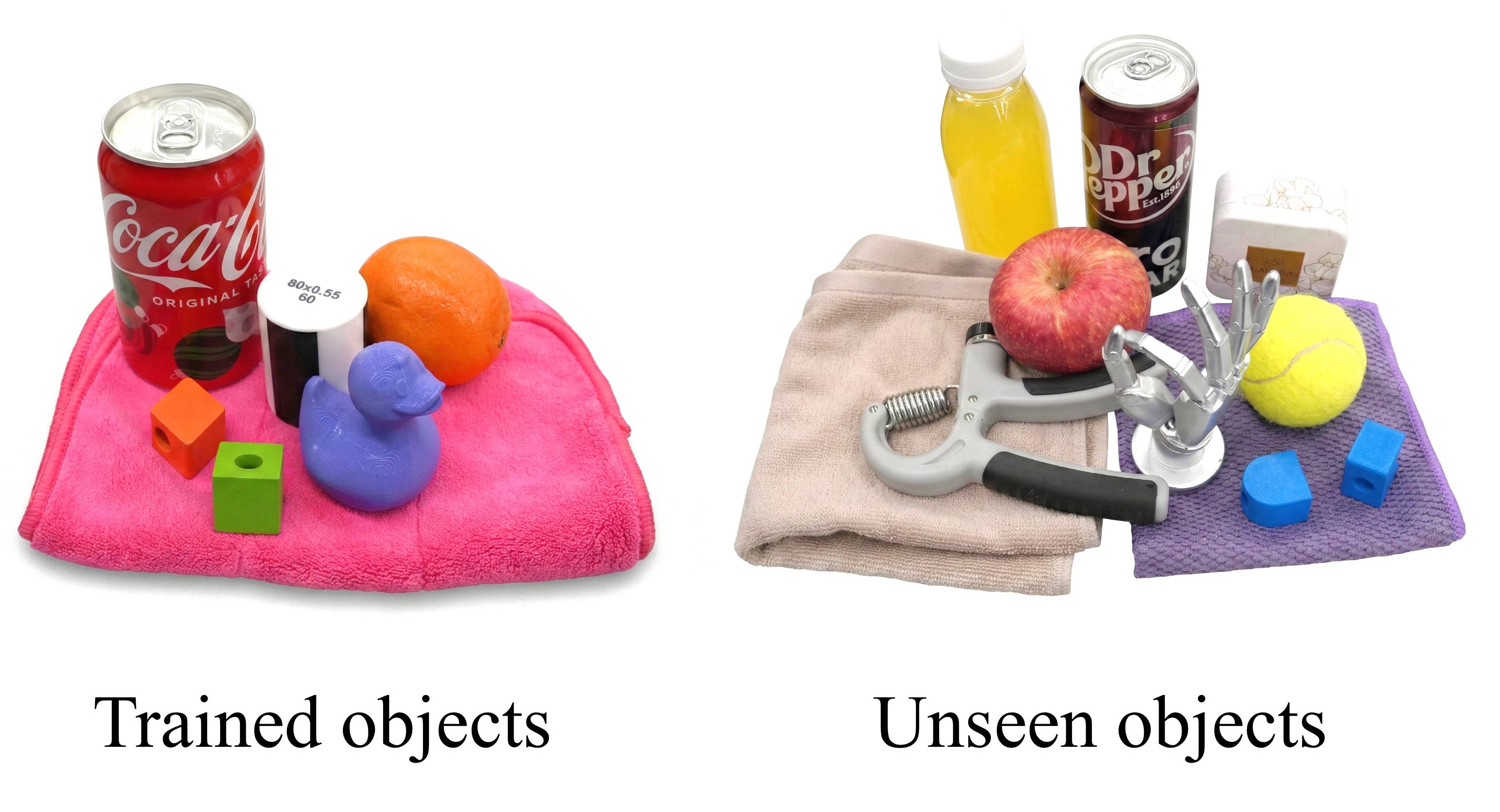}
  \caption{Trained and unseen object sets for evaluation. The collection encompasses diverse geometries with varying sizes, masses, and aspect ratios.}
  \label{fig:objects}
  \vspace{-2mm}
\end{figure}


 For the Grasp tasks, which involve relatively stable finger coordination, the model achieves a remarkably low mean absolute error (MAE) of 0.09 rad. In the more challenging In-hand Rotation tasks, characterized by complex joint coupling and rapid transitions, the validation error remains within a highly acceptable range of 0.15 rad. This minimal increase in error further validates the robustness of the EMG2Pose network, proving its capability to maintain high precision even during high-dimensional, dexterous manipulations.

\begin{figure*}[htb]
  \centering
  \includegraphics[width=\linewidth]{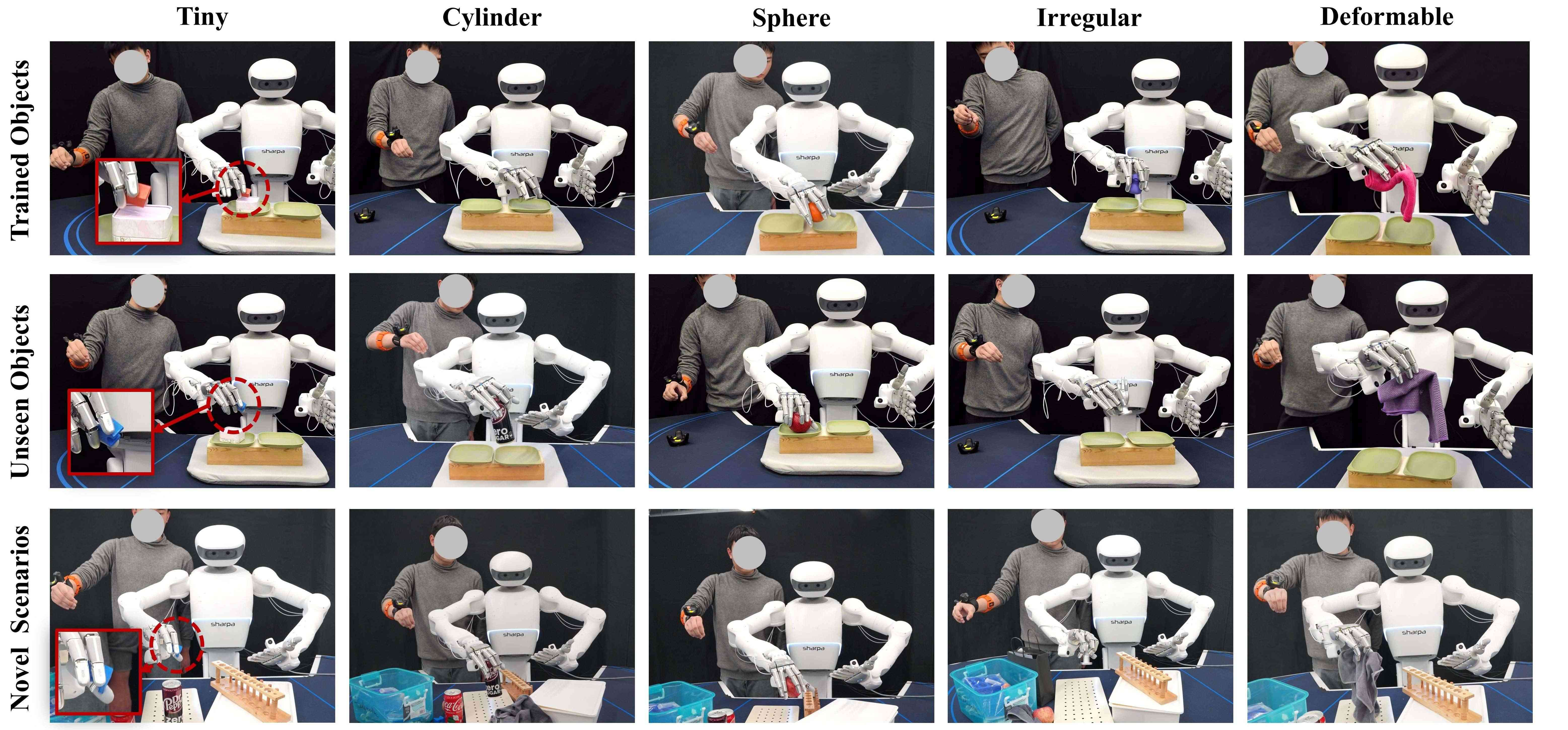}
  \caption{Grasping performance across diverse geometries and generalization scenarios, including trained objects, unseen objects, and novel environments.}
  \label{fig:grasp_matrix}
\end{figure*}

\begin{table*}[htb]
    \centering
    \caption{Detailed Performance Across Object Categories and Scenarios.}
    \label{tab:grasp_matrix}
    
    \small
    \setlength{\tabcolsep}{4.3pt}  
    \renewcommand{\arraystretch}{1.2}
    \begin{tabular}{l c c c c c c c c c c c c c c c c c}
        \toprule[1pt]
        \multirow{2}{*}{Scenario} & \multicolumn{2}{c}{Tiny} & & \multicolumn{2}{c}{Cylinder} & & \multicolumn{2}{c}{Sphere} & & \multicolumn{2}{c}{Irregular} & & \multicolumn{2}{c}{Deformable} & & \multicolumn{2}{c}{Overall} \\
        \cline{2-3} \cline{5-6} \cline{8-9} \cline{11-12} \cline{14-15} \cline{17-18}
        & SR$\uparrow$ & DR$\downarrow$ & & SR$\uparrow$ & DR$\downarrow$ & & SR$\uparrow$ & DR$\downarrow$ & & SR$\uparrow$ & DR$\downarrow$ & & SR$\uparrow$ & DR$\downarrow$ & & SR$\uparrow$ & DR$\downarrow$ \\
        \hline
        Trained Objects & 80.0\% & 18.8\% & & 90.0\% & 5.6\% & & 85.0\% & 5.9\% & & 50.0\% & 30.0\% & & 75.0\% & 20.0\% & & 76.0\% & 14.5\% \\
        Unseen Objects & 70.0\% & 14.3\% & & 75.0\% & 13.3\% & & 75.0\% & 13.3\% & & 40.0\% & 37.5\% & & 70.0\% & 21.4\% & & 66.0\% & 18.2\% \\
        Novel Scenarios & 45.0\% & 22.2\% & & 70.0\% & 14.3\% & & 80.0\% & 18.8\% & & 25.0\% & 60.0\% & & 60.0\% & 50.0\% & & 56.0\% & 28.6\% \\
        \bottomrule[1pt]
    \end{tabular}
    \vspace{-1mm}
\end{table*}

These results demonstrate that the EMG2Pose network is highly effective for standard grasping tasks and remains robust enough to interpret complex manipulation intentions. It provides a reliable kinematic foundation for the subsequent retargeting and real-time teleoperation.

\subsection{Generalization}

We curate two object sets (Fig.~\ref{fig:objects}), spanning five geometric categories. We further define a \textit{Novel Scenarios} condition, where object placements are randomized, and backgrounds are cluttered with occlusions. Each condition is tested over 20 trials per object category. We report two metrics: \textit{Success Rate (SR)}, the percentage of trials where the object is successfully grasped and placed at the target location, and \textit{Drop Rate (DR)}, the percentage of trials where a successful grasp results in the object being dropped. Representative grasping snapshots across all conditions are shown in Fig.~\ref{fig:grasp_matrix}, and quantitative results are summarized in Table~\ref{tab:grasp_matrix}. The system achieves an overall SR of 76.0\% with a DR of 14.5\% on the trained object set.

\textbf{Generalization to Unseen Objects.} When tested on novel objects within the same workspace, the overall SR decreases to 66.0\%, and the DR rises to 18.2\%. The performance degradation across categories remains moderate, indicating that the EMG2Pose model captures generalizable motor patterns rather than overfitting to specific object instances.

\textbf{Generalization to Novel Environments.} Under the Novel Scenarios condition, the overall SR further drops to 56.0\%. We attribute this degradation primarily to increased difficulty in arm-level planning: cluttered surfaces and randomized object placements introduce additional uncertainty during the approach phase, rather than indicating a fundamental failure of the EMG2Pose model itself. Supporting this interpretation, Sphere still achieves 80.0\% SR in the novel environment, as its symmetric geometry tolerates imprecise approach angles.

These results confirm that the DexEMG system provides multi-level generalization, from trained to unseen objects and from structured to unstructured environments, validating sEMG as a viable modality for portable dexterous teleoperation.
   
\begin{table}[t]
\centering
\caption{Sub-Phase Success Rates for Long-Horizon Manipulation Tasks.}
\label{tab:long_horizon_breakdown}
\small
\setlength{\tabcolsep}{4pt}  
\renewcommand{\arraystretch}{1.2}
\begin{tabular}{lcccc}
\toprule
Task & Grasp & Execution & \textbf{One-shot SR} & \textbf{With-retry SR} \\
\midrule
Packaging & 17/20 & 13/20 & 12/20 & 16/20 \\
Wiping    & 15/20 & 11/20 & 8/20 & 14/20 \\
\bottomrule
\end{tabular}
\vspace{-3mm}
\end{table}

\begin{figure*}[t]
  \centering
  \includegraphics[width=1\linewidth]{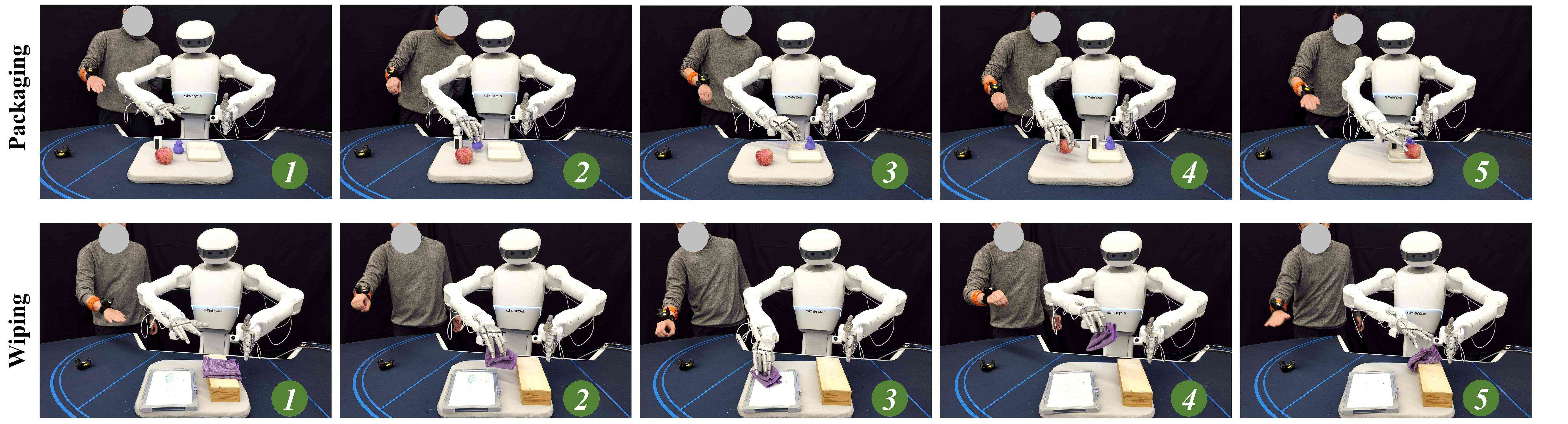}
  \caption{Long-horizon manipulation tasks: Sequential snapshots of Desktop Packaging and Table Wiping performed by the DexEMG system. These long-horizon tasks require several stages of manipulation and remain challenging.}
  \label{fig:long_horizon_tasks}
  \vspace{-1mm}
\end{figure*}

\subsection{Long-Horizon Manipulation and Scalability}

We select two representative long-horizon tasks (Fig.~\ref{fig:long_horizon_tasks}).
Both tasks require coordinated transitions between grasping, transporting, and executing contact-rich manipulations. Each task is performed for 20 trials. We decompose performance into sub-phase success rates: \textit{Grasp} (successful initial grasp), \textit{Execution} (completion of the subsequent manipulation), \textit{One-shot SR} (end-to-end success without re-attempts), and \textit{With-retry SR} (success when re-grasping is permitted).

As reported in Table~\ref{tab:long_horizon_breakdown}, the Packaging task achieves a One-shot SR of 60\%, while the more demanding Wiping task reaches 40\%. The lower Execution rate in Wiping stems from the need for sustained contact force during the sweeping motion, where even minor drift in the EMG-predicted pose can cause the cloth to slip. When the operator is permitted to re-attempt a failed grasp, the With-retry SR increases to 80\% for Packaging
and 70\% for Wiping, indicating that the system does not enter irrecoverable states after a failure.



Taken together, these results affirm that DexEMG can execute multi-stage, contact-rich tasks with reasonable success rates while remaining tolerant to intermittent failures. The lightweight setup and retry-friendly behavior make it a practical platform for scalable, dexterous data collection.

\section{DISCUSSION}
\textbf{Generalization via Neuromuscular Intent.}
DexEMG generalizes well to unseen objects and novel scenarios. Our model captures fundamental muscle synergy patterns. It avoids over-fitting to specific object geometries. This intent-based approach remains effective in cluttered environments. It is more robust than vision-based sensors under poor lighting conditions. The performance gap in irregular objects suggests a limitation. sEMG signals currently lack the signal-to-noise ratio for extreme tactile precision. High-precision tasks still require human-in-the-loop intervention.

\textbf{Latency and Stability Trade-off}
The input window size balances robustness and responsiveness. Longer temporal context filters muscle noise and stabilizes commands. However, this stability increases system latency. We choose a 400-step window for our experiments. This setting prioritizes reliability for tabletop manipulation.

\textbf{Limitations and Future Work.}
DexEMG currently requires individual calibration for new users. The system also lacks force feedback for handling fragile objects. Future work will explore cross-user foundation models to remove calibration needs. We also aim to integrate tactile feedback for high-precision manipulation.

\section{CONCLUSION}

We present DexEMG, a lightweight teleoperation system that uses a commodity sEMG armband to control a multi-fingered dexterous hand. By training an EMG2Pose network on synchronized muscle signals and hand poses, our system continuously predicts 22-DOF hand kinematics and maps them onto the robotic hand via a collision-aware retargeting algorithm. Experiments show that DexEMG achieves 76\% SR on trained objects and maintains reasonable performance on unseen objects and novel environments. The system also completes long-horizon tasks such as packaging and wiping with up to 80\% SR when retries are permitted. These results demonstrate that sEMG-based teleoperation offers a practical and scalable alternative to vision-based and exoskeleton-based systems for dexterous robotic manipulation.

\bibliographystyle{IEEEtran}
\bibliography{root}

@inproceedings{quivira2018translating,
  title={Translating sEMG signals to continuous hand poses using recurrent neural networks},
  author={Quivira, Fernando and Koike, Takahiro and Lukic, Milos and Sepulveda, Nakul and Montgomery, Mitchell and Shriraman, Arrvind and Menon, Carlo},
  booktitle={2018 7th IEEE International Conference on Biomedical Robotics and Biomechatronics (BioRob)},
  pages={166--171},
  year={2018},
  organization={IEEE}
}

@inproceedings{wen2021neuropose,
  title={NeuroPose: 3D Hand Pose Tracking using EMG Wearables},
  author={Wen, Shijia and Han, Jialiang and He, Fan and Zhang, Lin and Zhang, Yuntao and Ma, Siyuan and Xu, Yuanchao and Liu, Yunxin and Zhang, Li and Zhang, Mengmeng},
  booktitle={Proceedings of the Web Conference 2021 (WWW)},
  pages={3101--3112},
  year={2021}
}

@inproceedings{yi2021emg2pose,
  title={EMG2Pose: Estimating Finger and Wrist Pose from Surface Electromyography},
  author={Yi, Jennifer and Tang, Runzhe and Zhang, He and Chen, Xiaofeng and Belykh, Ivan and Taneja, Aarti and Salisbury, Curt and Saponas, T Scott and Kim, Sunghyun},
  booktitle={Advances in Neural Information Processing Systems (NeurIPS)},
  volume={34},
  pages={14543--14555},
  year={2021}
}

@inproceedings{emg2pose2024benchmark,
  title={EMG2Pose: A Large-scale Benchmark for sEMG-based Continuous Hand Pose Estimation},
  author={Chen, Xiaofeng and Zhang, He and Tang, Runzhe and Salisbury, Curt and Yi, Jennifer and Kim, Sunghyun and others},
  booktitle={Advances in Neural Information Processing Systems (NeurIPS)},
  year={2024}
}

@article{nieto2014teleoperation,
  title={Teleoperation of human-like robotic hands},
  author={Nieto, Jairo and Casals, Alícia and Rubio, Joan},
  journal={IEEE Robotics \& Automation Magazine},
  volume={21},
  number={4},
  pages={55--67},
  year={2014},
  publisher={IEEE}
}

@inproceedings{kessler1995cybergrasp,
  title={The CyberGrasp system},
  author={Kessler, G David and Hodges, Larry F and Walker, Neff},
  booktitle={Proceedings of the {V}irtual {R}eality {A}nnual {I}nternational {S}ymposium},
  pages={177--185},
  year={1995},
  organization={IEEE}
}

@article{sundaram2019learning,
  title={Learning the signatures of the human grasp using tactile gloves},
  author={Sundaram, Subramanian and Kellnhofer, Petr and Li, Yunzhu and Zhu, Jun-Yan and Torralba, Antonio and Matusik, Wojciech},
  journal={Nature},
  volume={569},
  number={7758},
  pages={698--702},
  year={2019},
  publisher={Nature Publishing Group}
}

@article{merriaux2017study,
  title={A study of {V}icon system acquisition reliability},
  author={Merriaux, Pierre and Dupuis, Yohan and Boutteau, Rémi and Vasseur, Pascal and Savatier, Xavier},
  journal={Sensors},
  volume={17},
  number={7},
  pages={1589},
  year={2017},
  publisher={MDPI}
}

@inproceedings{bassily2014robots,
  title={Robots and hand manipulation using contactless gestures},
  author={Bassily, Daniel and Georgoulas, George and Lytrivis, Panagiotis and Amditis, Angelos},
  booktitle={2014 IEEE International Conference on Robotics and Biomimetics (ROBIO)},
  pages={2122--2127},
  year={2014},
  organization={IEEE}
}

@article{lugaresi2019mediapipe,
  title={{M}edia{P}ipe: A framework for perceiving and processing multimodal data},
  author={Lugaresi, Camillo and Tang, Jiuqiang and Nash, Hadon and McClanahan, Chris and Uboweja, Esha and Hays, Michael and Zhang, Fan and Chang, Chuo-Ling and Yong, Ming and Lee, Juhyun and Kuang, Wan-Teh and Tan, James and Tseng, Matthias and Itti, Laurent and Grundmann, Matthias},
  journal={arXiv preprint arXiv:1906.08172},
  year={2019}
}

@inproceedings{supancic2015depth,
  title={Depth-based hand pose estimation: methods, data, and challenges},
  author={Supancic, James S and Rogez, Grégory and Yang, Yi and Shotton, Jamie and Ramanan, Deva},
  booktitle={Proceedings of the {IEEE} International Conference on Computer Vision (ICCV)},
  pages={1868--1876},
  year={2015}
}

@article{1_billard2019trends,
  author    = {Billard, Aude and Kragic, Danica},
  title     = {Trends and challenges in robot manipulation},
  journal   = {Science},
  volume    = {364},
  number    = {6446},
  pages     = {eaat8414},
  year      = {2019},
  publisher = {American Association for the Advancement of Science}
}

@inproceedings{3_daffertshofer2020teleoperation,
  author    = {Daffertshofer, Andreas and van den Heuvel, Margot R. and van der Sluis, Corry K. and Bongers, Raoul M.},
  title     = {Teleoperation of dexterous robotic hands: A review},
  booktitle = {Proceedings of the IEEE International Conference on Robotics and Automation (ICRA)},
  pages     = {10250--10256},
  year      = {2020}
}

@article{4_perez2017wearable,
  author    = {Perez-Gracia, Alba},
  title     = {Wearable exoskeleton glove for teleoperation of dexterous robotic hands},
  journal   = {IEEE/ASME Transactions on Mechatronics},
  volume    = {22},
  number    = {6},
  pages     = {2533--2544},
  year      = {2017}
}

@inproceedings{5_fang2020multimodal,
  author    = {Fang, Bin and Sun, Fuchun and Liu, Huaping and Guo, Di and Huang, Wenbing},
  title     = {Multimodal teleoperation interface for humanoid robot control},
  booktitle = {Proceedings of the IEEE/RSJ International Conference on Intelligent Robots and Systems (IROS)},
  pages     = {4033--4039},
  year      = {2020}
}

@article{6_sarac2019design,
  author    = {Sarac, Mine and Solazzi, Massimiliano and Frisoli, Antonio},
  title     = {Design and validation of a wearable exoskeleton for teleoperation},
  journal   = {IEEE Transactions on Haptics},
  volume    = {12},
  number    = {2},
  pages     = {132--144},
  year      = {2019}
}

@inproceedings{7_handa2020dextreme,
  author    = {Handa, Ankur and Van Wyk, Karl and Yang, Wei and Liang, Jacky and Chao, Yu-Wei and Wan, Qingtao and Birchfield, Stan and Ratliff, Nathan and Dieter, Fox},
  title     = {Dextreme: Transfer of agile in-hand manipulation from simulation to reality},
  booktitle = {Proceedings of the IEEE International Conference on Robotics and Automation (ICRA)},
  pages     = {5977--5984},
  year      = {2020}
}

@inproceedings{8_qin2022dexmv,
  author    = {Qin, Yuzhe and Wu, Yueh-Hua and Liu, Shaowei and Jiang, Hanwen and Yang, Ruihan and Fu, Yang and Wang, Xiaolong},
  title     = {DexMV: Imitation Learning for Dexterous Manipulation from Human Videos},
  booktitle = {Proceedings of the European Conference on Computer Vision (ECCV)},
  pages     = {570--587},
  year      = {2022}
}

@article{9_antoniou2021review,
  author    = {Antoniou, Zenonas and Dagalis, Thomas and Tachos, Nikolaos S. and Fotiadis, Dimitrios I.},
  title     = {A review of vision-based hand gesture recognition},
  journal   = {Journal of Imaging},
  volume    = {7},
  number    = {8},
  pages     = {133},
  year      = {2021}
}

@article{10_bi2019emg,
  author    = {Bi, Li-Zhong and Feleke, Adugna G. and Guan, Cuntai},
  title     = {A review on EMG-based control of assistive robotic devices},
  journal   = {IEEE Transactions on Neural Systems and Rehabilitation Engineering},
  volume    = {27},
  number    = {8},
  pages     = {1593--1605},
  year      = {2019}
}

@inproceedings{11_cai2021gesturedetect,
  author    = {Cai, Siyan and Chen, Yang and Huang, Shixuan and Fu, Yili and Liu, Honghai},
  title     = {GestureDetect: A real-time sEMG-based gesture recognition system},
  booktitle = {Proceedings of the IEEE International Conference on Robotics and Automation (ICRA)},
  pages     = {2361--2367},
  year      = {2021}
}

@article{12_zeng2021deep,
  author    = {Zeng, Wentao and Zhang, Sensen and Liu, Honghai and others},
  title     = {Deep Learning for Electromyography (EMG) Signal Analysis: A Review},
  journal   = {IEEE Access},
  volume    = {9},
  pages     = {156000--156015},
  year      = {2021}
}

@article{13_quivira2018translating,
  author    = {Quivira, Fernando and Koike, Takao and Marciano, Nicholas and others},
  title     = {Translating sEMG signals into continuous hand poses using deep learning},
  journal   = {IEEE Robotics and Automation Letters (RA-L)},
  volume    = {3},
  number    = {4},
  pages     = {3894--3901},
  year      = {2018}
}

@article{gu2016dexmo,
  title={Dexmo: An inexpensive and lightweight mechanical exoskeleton for capturing hand motion and providing haptic feedback in virtual reality},
  author={Gu, Xiaochi and Zhang, Yuru and Sun, Weibel and Bian, Yuanzhe and Zhou, Di and Lan, Pang},
  journal={IEEE Transactions on Visualization and Computer Graphics},
  volume={22},
  number={4},
  pages={1560--1569},
  year={2016},
  publisher={IEEE}
}

@inproceedings{pollard2002quantifying,
  title={Quantifying human hand ability},
  author={Pollard, Nancy S. and Gilbert, Todd M.},
  booktitle={Proceedings of the IEEE International Conference on Robotics and Automation (ICRA)},
  volume={4},
  pages={3540--3545},
  year={2002}
}

@inproceedings{fang2017pressure,
  title={Pressure-based hand gesture recognition using a flexible sensor glove},
  author={Fang, Bin and Sun, Fuchun and Liu, Huaping and Liu, Chao and Huang, Di and Guo, Di and Zhang, Tie},
  booktitle={Proceedings of the IEEE International Conference on Robotics and Biomimetics (ROBIO)},
  pages={2502--2507},
  year={2017}
}

@article{lin2021review,
  title={A review on wearable sensors for hand motion monitoring},
  author={Lin, J. and others}, 
  note={Note: This specific survey has a large author list, typically cited as Lin, J., et al.},
  author={Lin, Jiansheng and others},
  journal={Sensors},
  volume={21},
  number={10},
  pages={3415},
  year={2021},
  publisher={MDPI}
}

@inproceedings{han2016remote,
  title={Remote teleoperation of a humanoid robot using a vision-based gesture recognition interface},
  author={Han, Jung-Hyun and Kwak, Ji-Hye and Lee, Jung-Hoon and Park, Gwang-Hee and Lee, Sang-Hoon},
  booktitle={Proceedings of the IEEE International Conference on Systems, Man, and Cybernetics (SMC)},
  pages={1234--1239},
  year={2016}
}

@inproceedings{yuan2018depth,
  title={Depth-based {3D} hand pose estimation: From benchmark to challenges},
  author={Yuan, Shanxin and Garcia-Hernando, Guillermo and Stenger, Bjorn and Moon, Gyeongsik and Chang, Ju Yong and Lee, Kyoung Mu and Molchanov, Pavlo and Kautz, Jan and Honari, Sina and Ge, Liuhao and others},
  booktitle={Proceedings of the IEEE Conference on Computer Vision and Pattern Recognition (CVPR)},
  pages={1--16},
  year={2018}
}

\end{document}